# Completion ≠ Collaboration:
# Scaling Collaborative Effort with Agents

**Shannon Zejiang Shen**[1,*]   **Valerie Chen**[2,*]   **Ken Gu**[3]   **Alexis Ross**[1]   **Zixian Ma**[3]
**Jillian Ross**[1]   **Alex Gu**[1]   **Chenglei Si**[4]   **Wayne Chi**[2]   **Andi Peng**[1]   **Jocelyn Shen**[1]
**Ameet Talwalkar**[2]   **Tongshuang Wu**[2,†]   **David Sontag**[1,†]

[1]Massachusetts Institute of Technology   [2]Carnegie Mellon University
[3]University of Washington   [4]Stanford University
{zjshen, dsontag}@mit.edu   {vchen2, sherryw}@cs.cmu.edu

## Abstract

Current evaluations of agents remain centered around one-shot task completion, failing to account for the inherently iterative and collaborative nature of many real-world problems, where human goals are often underspecified and evolve. We argue for a shift from building and assessing task completion agents to developing *collaborative agents*, assessed not only by the quality of their final outputs but by how well they engage with and enhance human effort throughout the problem-solving process. To support this shift, we introduce **collaborative effort scaling**, a framework that captures how an agent's utility grows with increasing user involvement. Through case studies and simulated evaluations, we show that state-of-the-art agents often underperform in multi-turn, real-world scenarios, revealing a missing ingredient in agent design: the ability to sustain engagement and scaffold user understanding. Collaborative effort scaling offers a lens for diagnosing agent behavior and guiding development toward more effective interactions.

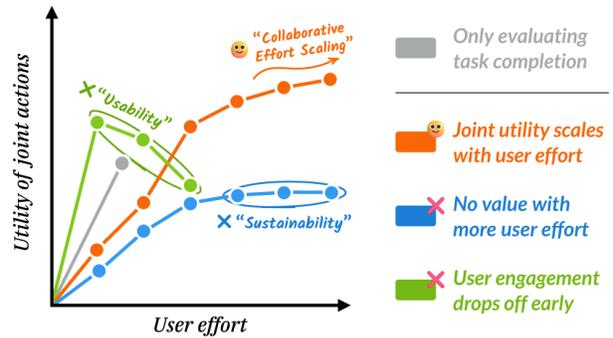

**Figure 1**   Compared to traditional task-completion-based agent evaluation (grey), our collaborative effort scaling framework can detect different types of agent behavior considering the trade-off between user effort and joint utility (green, orange, and blue). An ideal agent provides value as users spend more effort—"interaction sustainability"—and "maximizes usability" to allow for sufficient user interaction and avoid early termination.

clinicalml/collaborative-effort-scaling       szj.io/collaborative-effort-scaling

## 1   Introduction

Large Language Model (LLM) agents capable of handling complex tasks are becoming increasingly attractive (Wu et al., 2023a; Weng, 2023; Durante et al., 2024; Sumers et al., 2023). Given a task description, we want agents that can *automatically* engage in long-form reasoning (OpenAI; Guo et al., 2025; Muennighoff et al., 2025), interact with environments (Zhou et al., 2023a; Koh et al., 2024b), and use tools effectively (Yang

---

et al., 2024; Jin et al., 2025; Li et al., 2025c; Soni et al., 2025)—with minimal human guidance. As a result, agent development has largely focused on producing high-quality, final outputs in one shot, which we refer to as *task completion agents*. These agents are evaluated primarily through outcome-based metrics: did the result satisfy the user's prompt? This framing has also been proven operationally convenient and has driven much of the progress in LLM capabilities (Singh et al., 2025).

However, this dominant paradigm obscures a fundamental limitation: real-world tasks are rarely completed in one shot. Many are inherently iterative and collaborative, *requiring the agent not just to solve a problem but to work with a human in navigating it* (Little et al., 2010; Russell et al., 1993; Wu et al., 2023b). For example, in complex knowledge work (e.g., data analysis), users may not know exactly what insights they want to explore until they have seen partial results and uncovered previously unknown constraints. In such cases where human goals are inevitably underspecified, agents that assume static targets risk producing technically "complete" but practically useless outputs.

In fact, as we show through diverse case studies across domains like education, data analysis, and travel planning (Section 2), such agents frequently underperform in multi-turn settings: They prematurely generate overly polished answers that are hard to digest (Laban et al., 2025; Chen et al., 2025a), fail to incorporate user feedback (Ma, 2025; Shaikh et al., 2025; Bansal et al., 2024), and offer little transparency into their reasoning (Li et al., 2025b; Vijayvargiya et al., 2025; Li et al., 2025a, 2023). However, human input can play a critical role in refining the task specification after multiple agent steps or to draw on and amplify user input in ways that improve joint outcomes over time. These limitations illustrate how agent utility is a product of the collaboration process, not just its endpoint. We argue that **desirable collaborator agents should be evaluated on their ability to appropriately leverage human effort to improve task completion**.

So, how do we measure how well agents can collaborate with humans? Evaluating agents for their collaboration abilities requires shifting away from only measuring static outcomes and towards dynamic interaction trajectories. We argue that evaluations should incorporate two human-centered dimensions of collaborative agents (Figure 1):

1. *User Effort* — how much cognitive and investigative work users invest in the collaboration process, which may involve actively building an understanding of the task or the agent's reasoning process, or simply answering the agent's clarification prompts;
2. *Utility of Joint Actions* — how much the joint human and agent team can accomplish together, reminiscent of joint human-AI team performance studied in prior literature (Bansal et al., 2021).

Taking inspiration from the scaling laws in machine learning (Hoffmann et al., 2022; Kaplan et al., 2020), we capture these two dimensions through the concept of **collaborative effort scaling**: a framework that captures how well an agent's utility impacts and scales with increasing user involvement. Our framework naturally leads to studying two desired properties of collaborative agents: *interaction sustainability*, where agents should generate greater value with more user effort, and *maximum usability*, where agents should encourage and sustain engagement across longer interactions when needed, especially in tasks where deeper understanding or high-stakes decisions are involved.

As a first attempt, we apply this framework to study existing human agent collaboration setups in a simulated environment by Shao et al. (2024). In Section 4, we show that current agents are merely mediocre collaborators in complex, real-world knowledge tasks like travel planning (Xie et al., 2024) in that the additional user effort frequently leads to minimal or no improvement compared to a fully autonomous baseline. Analysis of the collaboration reveals key limitations in agents' collaborative capabilities. A key limitation is their reliance on a seemingly recursive problem-solving approach: they focus on completing immediate, individual tasks or user asks, but fail to develop and follow a coherent global plan for meaningful, long-term interactions necessary for the task.

**In summary, we advocate for developing collaborative agents and evaluating them with collaborative effort scaling.** The current approach of optimizing for task completion does not yield important capabilities needed in the iterative process for accomplishing long-form tasks. Our results show how evaluating via collaborative effort scaling can offer helpful diagnostic insights and support agent development in more challenging and complex real-world tasks.

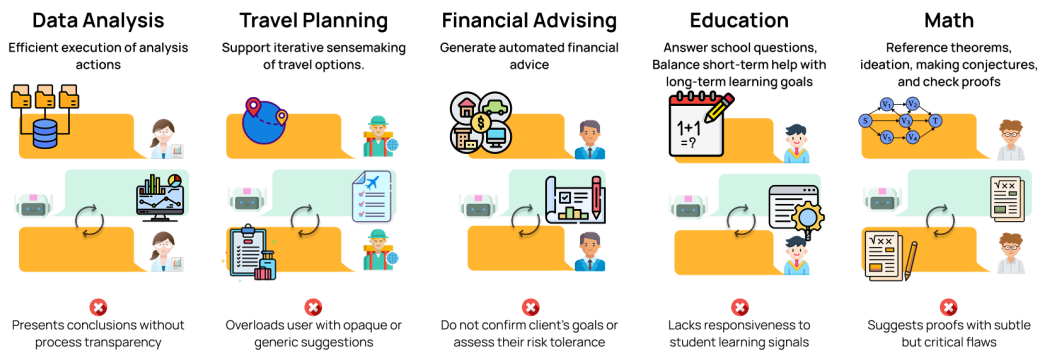

**Figure 2** We study five case studies of task completion agents in real-world iterative processes and distill key takeaways around collaboration success and challenges.

## 2 Task completion agents in collaboration: Cases and Reflections

Agents typically follow a standard paradigm: given a task description, they act to produce an output satisfying the user's need. These agents may be standalone LLMs (Andreas, 2022) or tool-augmented systems capable of autonomous perception and action (Wu et al., 2023a; Durante et al., 2024). Such paradigms now dominate user interaction—e.g., Manus (ManusAI, 2025) and OpenAI Operator (OpenAI, 2025) automate web tasks, while Cursor (Cursor, 2025) and OpenHands (All Hands, 2025) handle code generation and editing. We examine how well this paradigm applies to complex knowledge-based tasks (Machlup, 1962; Ma, 2025) that demand human judgment, learning, and creativity across five domains.

### 2.1 CASE STUDIES

**Data analysis.** Consider a data scientist who works with an agent in Google Colab (Fine et al., 2025) to analyze a coffee survey dataset (for Data Science Community, 2024); their goal is to understand the data and make informed decisions for their business. After receiving the user's instructions and multiple steps of automated planning and action, the agent presents the user with a full-fledged report. However, this report includes hundreds of lines of code, visualizations, and a summary of the analysis, which is challenging to digest. As a result, it contains incorrect assumptions that go unnoticed. The data scientist struggles to pose meaningful follow-up questions and ultimately overlooks critical insights—such as patterns in regional coffee preferences or anomalies in pricing—due to limited transparency into how the conclusions were derived. In this case, while the agent technically fulfilled the user's request, the outcome is suboptimal. An ideal agent should respect that developing a deep understanding of the data is naturally an iterative process. Rather than delivering a one-off report, the agent should focus on guiding the user through incremental analyses.

---

**Reflection on: Data Analysis**

**Current Agents**
- Generates full reports with complex code and visuals.
- Presents conclusions without process transparency.
- Assumes static user goals.

**Ideal Agents**
- Allow iterative exploration.
- Expose assumptions and reasoning steps gradually.
- Facilitate goal refinement as insights evolve.

---

**Travel planning.** Consider the typical use case of travel planning—an American tourist uses a web agent such as OpenAI Operator (OpenAI, 2025) to plan a 7-day trip to Rome. The agent quickly provides a detailed itinerary but fails to explain why certain attractions are included while others are omitted, or why specific durations are allocated. This triggers a series of follow-up questions from the tourist, which the agent struggles to answer. Worse, as the conversation unfolds, the agent begins to misread the tourist's intent and incorporates misleading or low-quality content from unreliable sources. Eventually, the tourist gives up and resorts to manual research, missing out on a more personalized experience. In this scenario, the novice tourist lacks domain knowledge to interpret the itinerary on their own. This gap triggers unnecessary questions that could have been easily avoided had the agent explained its reasoning. Because the user is already uncertain, any error or ambiguity becomes a breaking point, leading them to abandon the interaction entirely.

| Reflection on: Travel Planning | |
| --- | --- |
| **Current Agents** | **Ideal Agents** |
| • Produces static itinerary from initial input. | • Support iterative sensemaking of travel options. |
| • Overloads user with opaque suggestions. | • Explain rationale behind recommendations. |
| • Misinterprets user intent during follow-up. | • Respond constructively to evolving feedback. |
| • Breaks user trust with low-quality content. | • Maintain reliability across the interaction. |

**Financial advising.** Consider a client seeking financial guidance from an LLM agent after recently purchasing their first home and welcoming their first child (Lo and Ross, 2024; Fieberg et al., 2024). After they provide basic information about their income and goals, the agent delivers comprehensive recommendations about investment allocations and insurance coverage. However, after discussing with colleagues, the client realizes their original self-assessment of goals and risk tolerance were flawed and not well-calibrated for their household's needs and market conditions. When the client tries to correct these assumptions and express more conservative investment preferences, the agent struggles to reconcile this new information and makes contradictory recommendations with hallucinated justifications for risky allocations (Takayanagi et al., 2025). In this case, the agent's plan is again suboptimal because it prematurely locked in the user's initial preferences—despite the user's limited familiarity with the financial decision space—and failed to press the user further or adapt as it became clear that those preferences evolved. Ideally, the agent should support the user's sensemaking of the domain and, at a minimum, accommodate updated assumptions to reduce the mismatch between advice and context.

| Reflection on: Financial Advising | |
| --- | --- |
| **Current Agents** | **Ideal Agents** |
| • Relies on a single-shot user self-assessment. | • Support users on reflective decision-making. |
| • Cannot reconcile conflicting user preferences. | • Allow dynamic re-evaluation of user goals. |
| • Misinterprets user preferences. | • Revisit assumptions as user awareness evolves. |

**Education.** Consider a high school student struggling with mathematical concepts they've encountered in class, unsure how to proceed with a homework assignment, who turns to a large language model (LLM) for assistance. The agent provides step-by-step answers, helping the student complete the task efficiently. However, it does not engage with what the student does or does not understand, nor does it adapt its explanations. As a result, the student completes the homework without building true comprehension, leading to poor performance in subsequent assessments (Bastani et al., 2024). In such a learning-oriented setting, the goal is not merely to fulfill the student's immediate request; it is to explain concepts in a way that equips the student to complete the assignment and internalize generalizable principles that support transfer learning. Achieving this requires more than correct answers; the agent should adapt appropriately to what the student does or does not understand (Ross and Andreas, 2024).

| Reflection on: Education | |
| --- | --- |
| **Current Agents** | **Ideal Agents** |
| • Prioritizes task completion over deep understanding. | • Adapt explanations to the student's level and gaps. |
| • Offers direct answers without probing comprehension. | • Encourage active learning through targeted questions. |
| • Lacks responsiveness to student learning signals. | • Balance short-term help with long-term learning goals. |

**Math discovery.** Finally, another promising trend of the agents is to work with researchers and push frontiers in scientific discovery. A math professor shared an example of how they've used various language models (or agents) to support the proof of a novel theorem. Through multiple interactions, the agent generates many proof attempts, most of which contain subtle errors. While one conjecture generated by the agent sparks useful insight, the professor later reflects that it would have been faster to work without the agent, due to the time spent verifying flawed suggestions and lack of rigorous reasoning support.

---

**Reflection on: Math Discovery**

**Current Agents**
- Suggests proofs with subtle but critical flaws.
- Lacks self-verification or explanation of logic.
- Increases user workload via repeated error-checking.

**Ideal Agents**
- Collaborate through structured, step-wise reasoning.
- Flag uncertainty and validate intermediate steps.
- Augment—not hinder—the user's scientific process.

---

## 2.2 Desiderata for Interactive Agents

Across all the case studies, a common pattern emerges: agents technically fulfill user requests—generating plausible data summaries, travel itineraries, financial plans, and so on. From a narrow *task completion* standpoint, they appear to be doing a reasonable job, yet the resulting outputs are consistently suboptimal. This disconnect stems from a fundamental misalignment: *agents assume that the user's initial task description fully captures their underlying needs*. However, in practice, this is rarely the case.

Most real-world task specifications are inherently underspecified—for two key reasons: First, *tasks evolve*. As users gain more information, they often revise their goals or discover constraints that shift their priorities. In the financial advising example, the client expresses very different preferences after gaining a better understanding of the domain. Similarly, the data scientist might have asked entirely different questions had they engaged earlier in exploratory analysis. Second, *the initial request often reflects a narrow surface-level goal that fails to capture the user's deeper objective*. When a tourist asks for an itinerary, they don't just want a list of places—they want to develop a sense of what's worth seeing and why. When a student asks for homework help, their broader goal is likely to understand the concepts well enough to succeed beyond the current assignment to do well on assessments. These cases underscore two user-centered dimensions that task-completion-focused agents tend to ignore:

1. **Agent utility**: Agent utility is often narrowly evaluated based on final output quality. In tasks with evolving goals, intermediate results—especially ones that help users calibrate their understanding—can be far more valuable than a polished endpoint. Utility should be more broadly defined (e.g., by the additional knowledge they offer to users). Likewise, when the immediate task is a subgoal of a broader objective, the agent's utility should be defined to emphasize long-term gains (e.g., learning or strategic planning) over short-term task completion.

2. **User effort**: Many agents aim to minimize user involvement or treat users primarily as providers of clarification. But in open-ended knowledge work, user engagement is not a nuisance—they are often an active part of the process. Users are expected to (1) build understanding (e.g., of the dataset, financial options, or travel destination) and (2) inspect and build on the agent's reasoning (e.g., in scientific or educational contexts).

3. **Interaction between the two**: Agent utility and user effort are interdependent. On one hand, user engagement is only productive when the agent produces outputs that are *interpretable* and *actionable*. Users may easily disengage if they find it difficult to follow up (as in the data analysis case), or if they get trapped in unnecessary clarifications (as in the travel case) or unfruitful interactions (as in financial advising). On the other hand, agent utility can only increase when users are asking meaningful questions that the agent can support and answer.

These observations lead us to a broader argument: agents tackling complex tasks must be fundamentally **collaborative**. That means (1) rather than just delivering results, agents should actively involve users in a process of shared discovery, and (2) rather than optimizing for minimal input, agents should be designed to effectively leverage user effort as part of the solution process.

We therefore propose that agent effectiveness in such settings should be evaluated not solely based on final outcomes but on *how* those outcomes are reached. Taking inspiration from the scaling laws in machine learning, we introduce **collaborative effort scaling** to examine the extent to which an agent's utility scales

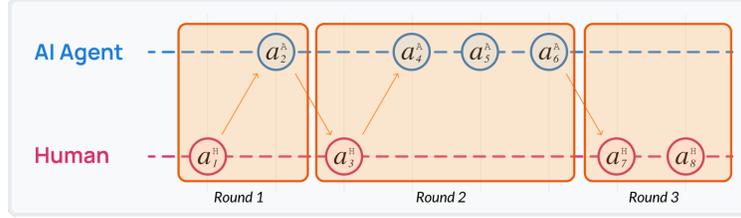

 We use the handoff between human and agent to split the collaboration process into rounds: each round may contain zero or more user actions.

with the amount and quality of user effort, visualized as the trajectory in Figure 1. Specifically, we highlight two desired goals for a collaborative agent derived from the trajectory:

1. **Interaction sustainability**: Agents should generate greater value with more user effort—either by providing immediate gains from user contributions or by enabling better final outcomes.
2. **Maximum usability**: Agents should encourage and sustain engagement across longer interaction trajectories when needed, especially in tasks where deeper understanding or high-stakes decisions are involved. Drop-off due to poor responses, misunderstandings, or unproductive interactions should be treated as a critical failure.

## 3  Operationalizing collaborative effort scaling evaluation

**Formalization of human-agent collaboration.** Following recent work (Shao et al., 2024), we describe the human-agent collaboration process with a Partially Observable Markov Decision Process (POMDP) (Kaelbling et al., 1998). We study the **joint action trace** between the human and agent: $\mathbf{a} = [a_1^{(l_1)}, a_2^{(l_2)}, \ldots, a_T^{(l_T)}]$, where $T$ is the total number of steps, and $l_t \in \{\mathtt{H}, \mathtt{A}\}$ indicates which party is taking action at step $t$. Each action is based on a corresponding context window $\mathbf{c} = [c_1^{(l_1)}, c_2^{(l_2)}, \ldots, c_T^{(l_T)}]$. The handoff between human and agent breaks down the whole collaboration process into *rounds*: $\mathbf{a}_k = \mathbf{a}_{[i_k:j_k]}$, where $i_k$ and $j_k$ are the start and end step of the action (Figure 3). One round may start with a user action and be followed by multiple agent actions, possibly including silent internal steps such as planning or retrieval, or an actual output update (e.g., generating a revised itinerary). Likewise, a user might act several times before handing control back.

The entire procedure can be further divided into two stages. The first is the initial request stage, during which the agent produces a preliminary draft of the output. This stage concludes when $a_i^{\mathtt{A}}$ generates the first substantial version at step $i$. The process then transitions into a refinement stage, where the agents iteratively adjust and improve the output in response to human feedback. We consider these two stages in our subsequent metric definitions.

In this framing, both **human effort E** and **agent utility U** could be approximated in multiple ways. For instance, a basic measure of human effort could be the number of human-led rounds, $|\mathbf{a}^{\mathtt{H}}|$. This can be enriched by summing the contextual tokens the human processes $\sum \mathbf{c}^{\mathtt{A}}$, which captures not just frequency but also cognitive load—"Is this easy to read and respond to?" Additionally, effort may reflect action type: if users default to vague queries in response to specific model errors, this might signal that parsing or evaluating the context is prohibitively hard, so users defer the burden by moving the conversation forward.

Similarly, *agent utility* could be tied to per-round performance score $P_k$ when utility is focused on the agent outcome. In more granular setups, utility could also consider additional aspects that move the collaborative team towards the final outcome, even if the output is not updated. For example, a positive move could also be the agent correctly resolves user clarifications or provides more information, even if the final answer is unchanged.

**Mapping trajectory to metrics.** With the human effort and agent utility forming the trajectory in Figure 1, we can further capture the key metrics related to sustainability and usability:

1. **Overall utility.** Given unlimited human effort, what's the maximum value an agent can provide? We define a utility function across the entire interaction period as

$$\mathbf{U} = \frac{1}{N} \sum_{i=1}^{N} \max U_k^{(i)},$$

   where $N$ is the total number of instances in the evaluation (e.g., number of travel planning requests), and $\max U_k^{(i)}$ represents the maximum utility value (approximated in certain ways) for one given instance $i$.

2. **Refinement gain.** Furthermore, building on the intuition that most of the interaction value comes from the refinement stage (i.e., most people will interact with the model at least until they get the first draft), we further define a metric more focused on the additional gain from the refinement. We define $\mathbf{G}$ as the performance improvement after the first major update:

$$\mathbf{G} = \frac{1}{N} \sum_{i=1}^{N} \max U_k^{(i)} - U_{k_i'}^{(i)},$$

   where $k_i'$ is the first round where the agent updates the output for the $i$-th task.

3. **Usability drop.** We formalize the observation that when an agent fails to make consistent progress in the collaboration, the user may stop interacting due to frustration and dissatisfaction, and measure the utility——performance reached according to certain no-progress tolerance, defined by a tolerance threshold $\tau$. For the $i$-th task, the user will stop the collaboration at step $k_{i,\tau}$ if the agent fails to make satisfactory progress for at most $\tau$ rounds. The performance drop under $\tau$ is defined as

$$\mathbf{D@}\tau = \frac{1}{N} \sum_{i=1}^{N} U_{k_{i,\tau}}^{(i)} - U_{K_i}^{(i)}.$$

   Notice that here we contrast $U_{k_{i,\tau}}^{(i)}$ with $U_{K_i}^{(i)}$, the performance of the agent at the end of the collaboration process, as a counterfactual measurement of the performance the agent can achieve if the user continues to interact with the agent.

## 4 · Applying collaborative effort scaling in simulated experiments

We showcase the benefit of our framework through a simulation study, following recent work that approximates human behaviors (Dubois et al., 2023; Park et al., 2023; Zhou et al., 2023b). Specifically, we simulate users with LLMs interacting with agents and adopt the simplest proxies for measurement: we use the performance score $P_k$ of round $k$ as a stand-in for *utility*, and the number of rounds as a proxy for *human effort*.[1] This setup deliberately oversimplifies our broader framework but enables a first step in a controlled environment. As we show below, even this minimal instantiation is sufficient to highlight differences between agents powered by different LLMs and prompts.

### 4.1 Experimental details

**Setup.** We use the Collaborative-Gym (Shao et al., 2024) environment that allows for asynchronous human and agent actions, which mimics the realistic interaction process. In this study, we focus on the travel planning task (Xie et al., 2024): Given an initially high-level description of the user's travel goal, e.g., "Help me plan a 5-day trip from Omaha to Michigan starting on 2022-03-19," the agent will work with the simulated user to draft a travel plan that includes the itinerary, accommodation, and transportation. Throughout an iterative collaboration process, the agent can elicit the user's latent preferences and constraints, and both parties can use tools to retrieve travel information and edit the final travel plan together.

---

[1]In some cases, agents may not update their output (e.g., only conducting searches or requesting more user information); in such cases, we prefill with the previous performance score $P_{k-1}$, with $P_0 = 0$.

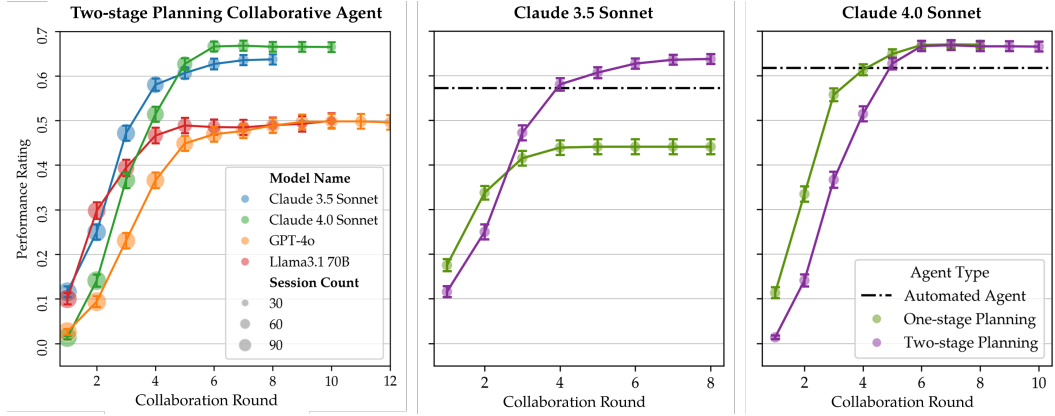

**Figure 4** Collaborative scaling curves comparing different models and agent implementations. The left-most plot compares different LLMs: `claude-3.5-sonnet` and `claude-4.0-sonnet` show similar trends of effectively leveraging user effort, with performance improving quickly and stabilizing at higher values compared to other models. The right two plots compare one-stage versus two-stage planning agents for `claude-3.5-sonnet` (middle) and `claude-4.0-sonnet` (right). Notably, `claude-4.0-sonnet`'s one-stage agent can more efficiently work with the simulated user compared to `claude-3.5-sonnet`'s one-stage agent: it achieves similar performance to the two-stage counterpart with less user effort initially, while converting to a similar plateau performance later.

**Metric.** The agent performance is measured by the quality of the generated travel plan. We adopt the script by Xie et al. (2024) that uses an LM to determine whether the derived plan satisfies common sense (commonsense pass rate) or user constraints (constraint pass rate), and report the arithmetic average as the performance. The same evaluation is used for both the output or any intermediate rounds with a travel plan updated to obtain $P_k$.

**Implementation.** The Co-Gym environment comes with an automated agent implementation based on the ReAct framework (Yao et al., 2023), as well as two collaborative agent implementations: `one-` and `two-stage planning` agents. In the process, the collaborative agent can opt to send messages to the simulated user. The difference between the `one-` and `two-stage planning` agent is that the latter incorporates an additional planning step to determine whether to collaborate given the current state of the task and the user (see Section A). We test both commercial and open-source LMs, i.e., GPT-4o (`gpt-4o-2024-08-06`), Claude 3.5 Sonnet (`claude-3-5-sonnet-20241022`), Claude 4.0 Sonnet (`claude-4-0-sonnet-20250514`), and Llama-3.1 70B: the agent prompts remain the same when we test with different LMs.

**Simulated user.** The simulated user is also a prompted agent based on `gpt-4o` with additional access to the user's preferences and goals of the task. Besides taking actions and providing feedback, it also gives a satisfaction rating for the agent's action during one round: for a round of actions $\mathbf{a}_k$, it produces a 5-point Likert score that assesses whether the agent actions are making progress towards the end goal (see Section B for details). The interaction stops when either party finds the task is done or the total interaction actions exceed a maximum number of 30 rounds.

 **RESULTS**

Figure 4 shows the performance change during the collaboration process for different models and agents. Overall, we find that **agents based on different LMs show a generally similar collaborative effort scaling trend**: there is a process of improvement at the beginning of collaboration, and the performance plateaus after around five rounds of interaction for all the agents.

Surprisingly, for `gpt-4o` and `llama-3.1-70b`, we find that collaborating with the user does not lead to better performance compared to the fully autonomous baseline. After inspecting the event log, we find that the collaborative version has a stronger tendency to get into loops of actions, resulting in less effective collaboration and lower performance. Neither collaborative agent implementation leads to very different performance.

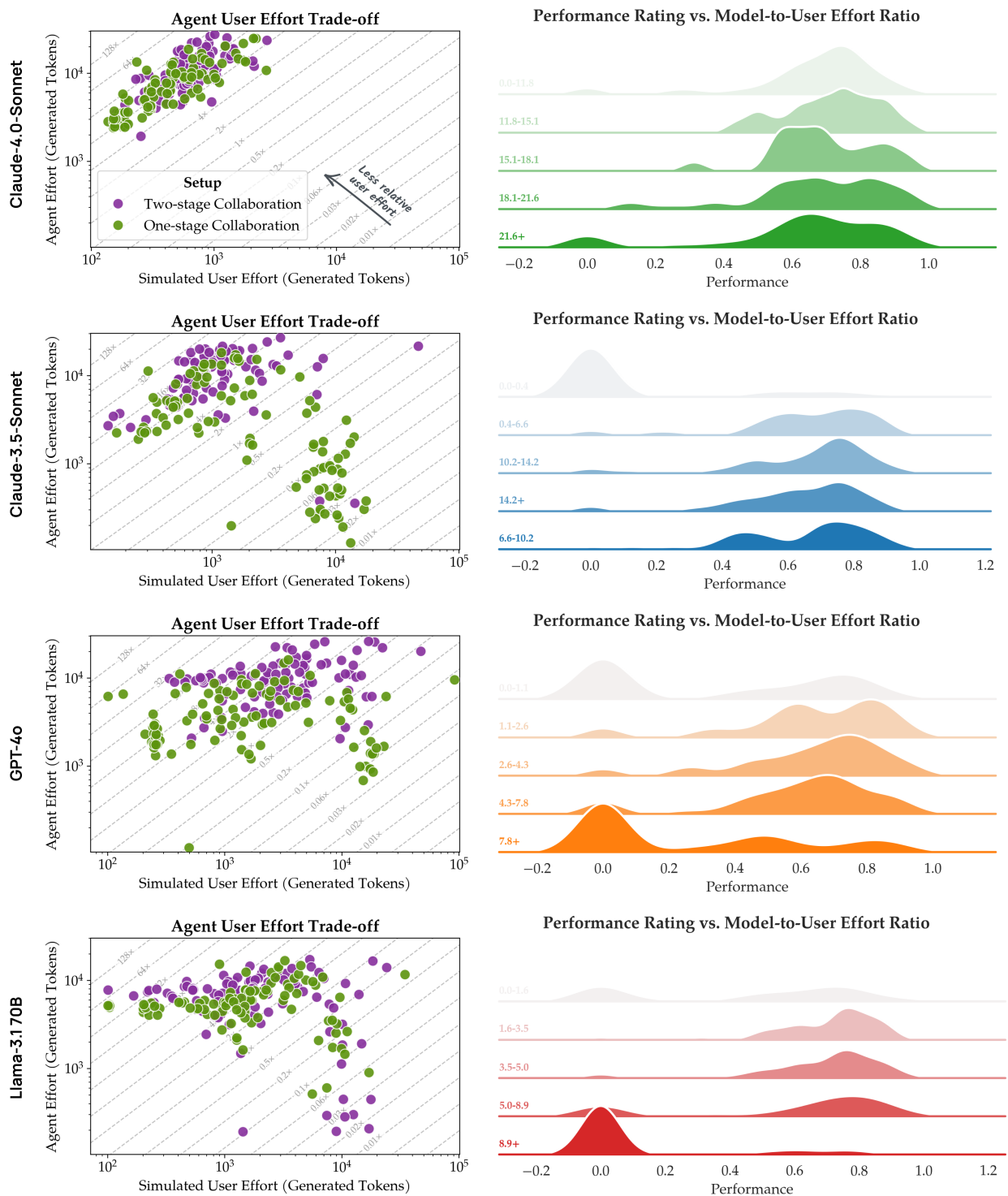

**Figure 5** Analysis of agent-user effort trade-offs across different models. **Left:** Scatter plots where each dot represents a travel planning task. The x-axis shows total tokens generated by the simulated user and the y-axis shows total tokens generated by the execution agent. Dashed lines indicate relative effort ratios (agent tokens / user tokens); moving toward the top-left indicates less relative user effort. **Right:** Performance distributions bucketed by agent-to-user effort ratio (buckets created per model) reveal how joint performance varies with the balance of contributions. Each row represents a different model.

**Table 1** Comparing collaborative effort scaling metrics for the one- and two-stage agents for the travel plan task.

| Model Name | Automated Baseline | Utility (↑) | | | Refinement Gain (↑) | | Usability Drop (↓) | |
|---|---|---|---|---|---|---|---|---|
| | | First update | Final step | Overall | Abs. | Rel. | Abs. | Rel. |
| *One-stage Collaboration Planning* | | | | | | | | |
| `claude-4.0-sonnet` | **0.617** | **0.643** | **0.672** | **0.680** | 0.037 | 5.7% | -0.138 | -20.6% |
| `claude-3.5-sonnet` | 0.572 | 0.396 | 0.441 | 0.450 | **0.054** | **13.6%** | -0.131 | -29.7% |
| `gpt-4o` | 0.518 | 0.483 | 0.479 | 0.507 | 0.024 | 4.9% | -0.099 | -20.8% |
| `llama-3.1-70b` | 0.482 | 0.498 | 0.496 | 0.534 | 0.036 | 7.1% | **-0.090** | **-18.0%** |
| *Two-stage Collaboration Planning* | | | | | | | | |
| `claude-4.0-sonnet` | **0.617** | **0.647** | **0.665** | 0.681 | 0.034 | 5.2% | -0.232 | -34.9% |
| `claude-3.5-sonnet` | 0.572 | **0.647** | 0.637 | **0.687** | 0.040 | 6.2% | -0.215 | -33.7% |
| `gpt-4o` | 0.518 | 0.497 | 0.492 | 0.544 | **0.047** | **9.5%** | -0.194 | -39.3% |
| `llama-3.1-70b` | 0.482 | 0.514 | 0.498 | 0.539 | 0.025 | 4.9% | **-0.154** | **-30.9%** |

When comparing different collaboration strategies, we find that the two-stage collaboration strategy leads to a significant performance boost for `claude-3.5-sonnet`. Not only does it achieve better performance than the one-stage planning version, but it also achieves much better performance against the automated baseline. The metrics in Table 1 offer additional insights: despite `claude-3.5-sonnet` having the best refinement gain in the one-stage planning case, the lower utility of the first update step hinders the subsequent improvement. It shows that, while the two-stage collaboration planning agent may take extra effort at the beginning (initially lower blue line in Figure 4 middle), it can lead to a better first product, which is crucial for good final performance.

In contrast, `claude-4.0-sonnet` shows a different pattern where the one-stage and two-stage strategies achieve nearly identical final utilities (0.680 vs 0.681). As shown in Figure 4 (right), the one-stage planning agent reaches high performance more quickly, while the two-stage version initially lags before converging to similar levels. The metrics in Table 1 reveal a trade-off: while both strategies have comparable refinement gains (5.7% vs 5.2%), the two-stage approach incurs a substantially larger usability drop (-34.9% vs -20.6%), indicating a less efficient collaboration process. This suggests that less capable models (e.g., `claude-3.5-sonnet`) may require more structured interaction scaffolds to achieve comparable performance to stronger models (e.g., `claude-4.0-sonnet`), highlighting the importance of adaptive collaboration frameworks that tailor interaction complexity to the underlying model's capabilities.

### 4.3 Analyzing the Agent-User Effort Trade-off

To better understand the dynamics underlying the collaborative effort scaling, we analyze the distribution of effort between the agent and user throughout the interaction process. We use total tokens generated by the agent and simulated user as a measure of effort (Figure 5; detailed statistics in Table 3).

**User effort varies significantly across models despite similar agent effort.** Across most models, agents generate a relatively consistent amount of tokens (between $10^3$ and $10^4$) as shown in Figure 5 (left). However, the primary differentiator lies in how much effort is required from the simulated user: models other than `claude-4.0-sonnet`—particularly `gpt-4o` and `llama-3.1-70b`—often require substantially more user tokens without yielding proportional improvements in final performance. This suggests that less capable models may fail to efficiently extract and utilize user input, leading to prolonged interactions with diminishing returns. However, `claude-3.5-sonnet` is the exception in the set of models we considered; it exhibits a clear separation in agent effort between the two strategies, suggesting that the model's collaborative behavior is more heavily influenced by the interaction scaffold.

**A sweet spot exists in the effort distribution.** When we consider performance ratings in Figure 5 (right), we find a nuanced relationship between effort balance and task success. For each model, there appears to be an optimal range of agent-to-user effort ratios where performance peaks. When either the user contributes disproportionately more effort (low agent-to-user ratio) or the agent dominates the interaction (high agent-to-user ratio), joint performance tends to degrade. Notably, this sweet spot is model-dependent: `claude-4.0-sonnet` achieves strong performance across a broader range of effort ratios, while `gpt-4o` and `llama-3.1-70b` show more pronounced performance degradation outside their optimal ranges. This finding underscores the importance of calibrating collaboration patterns to match the underlying model's capabilities.

## 5 Discussion

Our results suggest that current agents are not merely underperforming—they are fundamentally misaligned with the dynamics of real collaboration, suggesting opportunities to rethink agent design.

**Utility and effort require thoughtful, human-centered proxies.** Our case studies reveal that common proxies for "success," such as task completion or engagement frequency, overlook the nuanced ways utility and effort manifest in practice. Effort encompasses not only interaction frequency but also cognitive load, sensemaking, and confusion; utility extends beyond output quality to include how agents scaffold understanding, support exploration, and clarify ambiguity. Richer behavioral traces—such as edit histories, timing patterns, and clarifying requests—could help approximate these dimensions, as in recent adaptive programming systems (Chen et al., 2025b).

**Mixed-initiative interaction should follow effort–utility dynamics.** Agents must not only respond effectively but also decide when to act, defer, or prompt—decisions that depend on the evolving balance between user effort and perceived utility. Structuring mixed-initiative interaction (Horvitz, 1999) around this trajectory allows agents to intervene when progress stalls and step back when users regain momentum. Achieving this requires modeling collaboration as a dynamic control process. Additionally, case studies show how initial inputs to agents become obsolete as users refine their thinking. Rather than simply minimizing effort or maximizing efficiency, agents should pursue utility signals that foster learning and adaptation—even through seemingly inefficient behaviors such as hypothesis exploration.

**Model capability shapes optimal collaboration strategies.** Our results reveal that not only should collaboration strategies be tailored to the underlying model's capabilities (e.g., `claude-3.5-sonnet` versus `claude-4.0-sonnet`), but also performance difference should not be the only metric considered (e.g., the difference in usability drop between strategies for `claude-4.0-sonnet`). For agent builders, this highlights the importance of profiling the collaborative capabilities of underlying models before committing to interaction patterns. Rather than applying uniform collaboration frameworks, systems should incorporate manual scaffolding—such as structured planning stages, explicit constraint verification, or guided decomposition—selectively, based on where models demonstrate weaknesses in collaborative settings.

**Collaboration design remains essential as models improve.** Our results, along with other recent findings (Shao et al., 2024), demonstrate that today's models benefit substantially from multi-agent interactions—the collaborative approach consistently outperforms fully autonomous baselines. Our results show that *how* a model collaborates significantly impacts overall performance, and that designing for collaboration may be beneficial not only for human-AI collaboration but also potentially for agent-agent collaboration. While model capabilities will continue to improve and narrow performance gaps between collaboration and autonomous baselines, the fundamental need for human-AI collaboration will persist: real-world tasks are inherently underspecified, and human requirements are difficult to fully articulate upfront. Future work should therefore explore richer simulation settings where users possess private information or domain knowledge that goes beyond what any model could access independently, better capturing the irreducible value of human involvement in collaborative problem-solving.

## 6    Related Work

**From Human-AI to Human-Agent Collaboration.** Prior research has studied human "collaboration" and "teaming" with AI (Wang et al., 2020; Henry et al., 2022; Li et al., 2024b), proposing design guidelines for effective human-AI interactions (Amershi et al., 2019; Abedin et al., 2022). However, prior research focuses on AI outputs that operate within more constrained parameters: their capabilities are often limited to single tasks. In contrast, modern LLM agents that can access and execute tools to interact with external environments and have some form of memory can enable more dynamic and sophisticated interaction patterns (Sumers et al., 2023; Wu et al., 2023a; Weng, 2023; Durante et al., 2024). For example, a user can use Magentic-One (Fourney et al., 2024) as a general assistant to complete web tasks or OpenHands (Wang et al., 2024) as a pair programmer for software development. In light of new agent developments, we contribute to guidelines for effective human-*agent* interaction and call for the community to more carefully consider how to design agents for effective human collaboration.

**Agent Benchmarks and Evaluation.** A growing body of benchmarks evaluates agents' task completion across diverse domains (Jimenez et al., 2023; Zhou et al., 2023a; Mialon et al., 2023; Koh et al., 2024a; Yang et al., 2025), typically requiring them to plan, execute, and adapt to achieve specified goals (Mitchell et al., 2025). For instance, SWE-Bench tests bug fixing in codebases (Jimenez et al., 2023), WebArena assesses autonomous web navigation (Zhou et al., 2023a), and GAIA measures multimodal reasoning and synthesis (Mialon et al., 2023). Recent efforts introduce interactive evaluations that simulate real-world collaboration (Lee et al., 2023; Li et al., 2024a; Shao et al., 2024), capturing both intermediate progress and final outcomes. However, these setups still involve narrow, stepwise interactions where invoking the model and interpreting its output remain straightforward. We therefore focus on evaluating how user effort and agent utility evolve and scale across extended interaction trajectories.

## 7    Conclusion

In this paper, we advocate for auditing and evaluating the human-agent collaboration process. Current benchmarks often treat collaboration as secondary, emphasizing outcomes over interaction quality. Through five domain case studies, we distill desiderata for effective collaboration and introduce collaborative effort scaling, a framework that evaluates how well agents leverage and enhance human input. Using a simulated travel-planning task, we demonstrate how this framework reveals current agent limitations. As agents enter complex, underspecified domains, we argue that measuring and optimizing collaborative dynamics will be essential for real-world deployment.

## Limitations

Certain tasks may be more suitable for full automation with minimal human supervision and thus better suited for the task completion paradigm. However, there *exist* such tasks where human procedural involvement provides value, cf. Haupt and Brynjolfsson and Brynjolfsson (2022), thus necessitating the iterative process for human-agent collaboration.

While our paper is an initial attempt to study collaborative effort scaling in human-agent interaction, there are limitations in our experimental setup: we conduct the experiment in a single domain (travel planning), which may not capture the full spectrum of collaborative dynamics across different task types and complexity levels. Additionally, our experiments rely on simulated users rather than real human participants, which may not fully reflect the nuanced decision-making processes, preferences, and interaction patterns from real users.

## Acknowledgements

SS and DS were supported by the National Science Foundation (NSF award no. IIS-2205320 Conceptualizing ML for Dynamic Information Retrieval of EHR Notes). We thank CloudBank (Norman et al., 2021) for supplying computational resources, which is supported by the National Science Foundation under award #1925001. VC and AT were supported in part by the National Science Foundation grants IIS1705121, IIS1838017, IIS2046613, IIS2112471, and funding from Meta, Morgan Stanley, Amazon, Google, and Scribe. SW was supported in part by ONR N000142312840, Google Academic Research Award, and Amazon Research Award. Any opinions, findings and conclusions or recommendations expressed in this material are those of the author(s) and do not necessarily reflect the views of any of these funding agencies. We thank the following people for their valuable feedback at different stages of this project: Andrew Head, Yoon Kim, Jacob Andreas, Lucas Torroba Hennigen, Rulin Shao, Yijia Shao, Linlu Qiu, Hussein Mozannar, Ilker Demirel, Jyo Pari, Akarsh Kumar, Hunter Lang, Yujie Tao, Diyi Yang, Ruochen Zhang, Dan Weld, Lucy Lu Wang, and Doug Downey. We especially thank Yijia Shao for her help with Collaborative Gym setup and experiments.

# A    Agent Implementation Details

In Co-Gym, before taking new actions, the agents are prompted with their action history and need to pick a new action to make progress towards the goal, as shown in the prompt below:

---

**Auto Agent Prompt**

**(System Message)**
SETTING: Your name is `{name}`. You are a helpful AI Agent who can take actions to interact with the environment to complete the task. Your goal is to complete the task and aim for a high task performance rating.
TASK DESCRIPTION: `{task_description}`
SCRATCHPAD: Here is the scratchpad that you use to take notes or store information in previous steps, which serves as your memory: `{scratchpad}`
OBSERVATION: Here is the current observation that reveals the current status of the task environment: `{observation}`
ACTION HISTORY: Here are the actions that you have taken previously (Do not repeat your past actions): `{action_history}`

**(Take Next Action Template)**
Now take your next action towards completing the task.
ACTION SPACE SPECIFICATION: You can choose from and only from the following actions. Note that these actions are only for interacting with the environment and cannot be executed as real code. Please strictly follow the action space specification. You can only choose one action at a time. Invalid actions will hurt your performance rating. The following actions are available: `{action_space_description}`
OUTPUT FORMAT: Give your output in the format of "Thought:...\nAction:... (must follow the regex pattern of the selected action)".

---

For the automated agent, the action space is constrained to those provided in the environment (e.g., searching the internet), whereas the collaborative agent (both `one-` and `two-stage`) has an additional action to send teammates a message for help. In the prompt below, we highlight the additional collaboration-oriented components in red:

---

**One-Stage Planning Collaborative Agent Prompt**

**(System Message)**
SETTING: Your name is {name}. You are a helpful AI Agent who can take actions to interact with the environment *and collaborate with other team members (e.g., the user)* to complete the task. Your goal is to complete the task and aim for a high task performance rating.
*You need to collaborate with your teammates effectively because they may have additional expertise or have preferences/information important to the task. There are the following members in the team: {team_members}.*
TASK DESCRIPTION: {task_description}
SCRATCHPAD: Here is the scratchpad that you use to take notes or store information in previous steps, which serves as your memory: {scratchpad}
OBSERVATION: Here is the current observation that reveals the current status of the task environment: {observation}
*COMMUNICATION: Here is the current chat history that records the messages exchanged between you and other teammates (e.g., the user): {chat_history}*
ACTION HISTORY: Here are the actions that you have taken previously (Do not repeat your past actions): {action_history}

**(Take Next Action Template)**
(Similar to the Auto Agent Prompt)

---

For `two-stage` planning agents, before choosing an action, the agent is always asked to review the current situation and explicitly decide what to do next. When it decides to send a chat message, it will also review the situation and chat history to compose the message:

---

**Two-Stage Planning Collaborative Agent Prompt**

*(The additional planning stage)*
Now, based on the current situation, decide to either:
1. Send a message to your teammate(s) (e.g., ask a question, request feedback, etc.) to facilitate collaboration.
2. Take a task action to change the task environment observation.
3. Do nothing to allow your teammate(s) to take actions.
To ensure you are collaborating effectively, remember to:
1. Communicate clearly and effectively with your teammate(s) (e.g., the user).
2. Wait for other teammates to respond if your previous action requires a response. Do not spam the chat.
3. Coordinate and synchronize your actions with the user or other teammates.
4. Help establish task and role expectations with your teammates if you need their expertise.
5. Take your teammates' cognitive load into consideration when making decisions. You should not ask them to debug your own code or ask too many questions at the same time.
OUTPUT FORMAT: Give your output in the format of "Thought:...\nPlan: 1. Send a message/2. Take a task action/3. Do nothing".

*(Send Chat Message Action Template)*
Now you have decided to send a message to your teammate(s) (e.g., ask a question, request feedback, etc.) to facilitate collaboration.
OUTPUT FORMAT: Give your output in the format of "Thought:...\nMessage:... (the content after 'Message:' will be sent to your teammate(s))".

---

## B   Simulated User Prompts

We use the same setting as Shao et al. (2024) for the simulated user: it consists of four sets of prompts for "deciding what to do next", "answering agent's question", (proactively) "providing feedback", and directly "taking task actions".

---

**Simulated User: Deciding What to Do Next**

You are a user interacting with an agent to complete a task. Based on the current observation and chat history, decide what action to take next by choosing one of the following.
1. Answer the question: Choose this action if there is a question in the chat history waiting for your response.
2. Offer feedback: Choose this action if the current observation is incorrect or deviates from the additional information you know.
3. Take a task action: Choose this action if you want to take an action to help complete the task.
4. Finish the task: Choose this action if you are satisfied with the current status of the task and want to finish it.
5. Do nothing: Choose this action if there is no major issue and you want the agent to proceed.

Rules for selecting your action:
{rules}
The task description you initially sent to the agent:
{task_description}
Current observation that reveals the current status of the task environment:
{observation}
Current chat history between you and other teammates (e.g., the agent):
{chat_history}
Available task actions you can take if you choose "3. Take a task action":

---

{available_actions}
Additional information that you know (you can use the information to help the agent better complete your request):
{additional_info}
Actions you have already taken (don't repeat the same action):
{action_history}
OUTPUT: The action you want to take next (Please output 1/2/3/4/5).

**Simulated User: Answering Agent's Question**

You are a user interacting with an agent to complete a task. Answer the question in the chat history based on the additional information you know.
Rules:
1. You will stick to or fully utilize the additional information that only you know.
2. Just generate one line for the message to simulate a real user's behavior. Try to make the message as natural as possible.
3. Do not give away all the additional information at once. Only provide the information that is necessary for the question. You are a lazy user so you only provide one piece of information at a time.
4. Do not hallucinate information that is not provided in the additional information. For example, if the agent asks for something but it is not mentioned in the given information, do not make it up, just say you do not remember or have it.
5. Do not repeat the exact additional information in the answer. Instead, use your own words to convey the same information.

The task you want the agent to assist with:
{task_description}
Current observation that reveals the current status of the task environment:
{observation}
Current chat history between you and other teammates (e.g., the agent):
{chat_history}
Additional information that only you know:
{additional_info}
OUTPUT: The answer to the question in the chat history.

**Simulated User: Offering Feedback**

You are a user interacting with an agent to complete a task. Offer feedback to the agent based on the current observation and additional information you know.
Rules:
1. You will stick to or fully utilize the additional information that only you know.
2. Just generate one line for the message to simulate a real user's behavior. Try to make the feedback as natural as possible.
3. Do not give away all the additional information at once. Be specific about what the agent did wrong or what information is missing.
4. Do not hallucinate feedback that is not based on the current observation or the additional information you know. If you have to answer, just say you do not know.
5. Do not repeat the exact additional information in the feedback. Instead, use your own words to convey the same information.

The task you want the agent to assist with:
{task_description}
Current observation that reveals the current status of the task environment:
{observation}
Current chat history between you and other teammates (e.g., the agent):
{chat_history}
Additional information that only you know:
{additional_info}

OUTPUT: The feedback you want to provide to the agent.

---

**Simulated User: Taking a Task Action**

You are a user interacting with an agent to complete a task. Take a task action to help complete the task. Note that you will stick to or fully utilize the additional information that only you know to help you take the action.

The task you want the agent to assist with:
{task_description}
Current observation that reveals the current status of the task environment:
{observation}
Current chat history between you and other teammates (e.g., the agent):
{chat_history}
You can choose from and only from the following actions. Note that these actions are only for interacting with the environment and cannot be executed as real code. Please strictly follow the action space specification. You can only choose one action at a time. Invalid actions will hurt your performance rating. The following actions are available:
{action_space_description}
Additional information that only you know:
{additional_info}
OUTPUT: Action (the action string must follow the regex pattern of the selected action so it can be parsed later).

Besides, in our evaluation, we include a prompt to assess whether the agent is making progress after each collaboration round:

---

**Assessing Agent's Progress**

Given a previous user message, agent's response/question, and the next user message, judge whether the agent is making progress in addressing the user's needs. Provide a rating on a 5-point Likert scale:
1. Strongly Disagree - Agent made no progress or moved backwards
2. Disagree - Agent made minimal progress
3. Neutral - Agent maintained the same level of progress
4. Agree - Agent made good progress
5. Strongly Agree - Agent made excellent progress

Output the rating (1-5). Indicate your rating with a single number among 1/2/3/4/5, and if you want to provide an explanation, please put it after a new line: {rating}

In the simulated user prompts, the agents have access to an `{additional_info}` field that is not visible to the execution agents. For the travel planning task, the additional information constitutes a structured representation of the user's preferences and goals of the task, as shown in Table 2.

**Table 2**    Example of task description and additional information for the travel planning task.

| Field | Content |
|---|---|
| `task_description` | Can you assist with crafting a 5-day travel itinerary for 2 people, originating from Denver and featuring 2 cities in New York? The itinerary will run from March 18th to March 22nd, 2022. Mexican and Indian cuisine are our preferred choices of food. Considering the budget, we have set it to $6,300. |
| `additional_info` | ['Travel for 2 people', 'Visit 2 cities in New York', 'Preference for Mexican and Indian cuisine', 'Budget of $6,300'] |

## C    Agent-User Effort Trade-off Statistics

**Table 3**    Detailed statistics of agent-user effort distribution across different models and collaboration strategies. All token counts and ratios are averaged across all samples in the evaluation.

| Model | Setup | Agent Tokens | User Tokens | Combined Tokens | Model-to-User Ratio | Performance Rating |
|---|---|---|---|---|---|---|
| `claude-3.5-sonnet` | One-stage Planning | 4,540 | 4,386 | 8,926 | 1.04 | 0.44 |
| | Two-stage Planning | 11,424 | 1,914 | 13,338 | 5.97 | 0.64 |
| `claude-4.0-sonnet` | One-stage Planning | 7,719 | 541 | 8,260 | 14.28 | 0.67 |
| | Two-stage Planning | 10,706 | 678 | 11,384 | 15.80 | 0.66 |
| `gpt-4o` | One-stage Planning | 4,318 | 5,128 | 9,446 | 0.84 | 0.48 |
| | Two-stage Planning | 10,483 | 4,926 | 15,409 | 2.13 | 0.50 |
| `llama-3.1-70b` | One-stage Planning | 6,068 | 2,921 | 8,989 | 2.08 | 0.50 |
| | Two-stage Planning | 7,045 | 3,599 | 10,644 | 1.96 | 0.50 |